\documentclass{ifacconf}

\usepackage{graphicx}      
\usepackage{natbib}        

\makeatletter
\let\old@ssect\@ssect 
\makeatother

\usepackage{graphics} 
\usepackage{epsfig} 
\usepackage{amsmath} 
\usepackage{amssymb}  
\usepackage{enumerate}
\usepackage{float} 
\usepackage[linesnumbered,ruled,algo2e]{algorithm2e} 
\usepackage{adjustbox}
\usepackage{subcaption} 
\usepackage{hyperref}
\usepackage[]{units}
\usepackage{bm}

\usepackage{array}
\newcolumntype{P}[1]{>{\centering\arraybackslash}p{#1}}

\usepackage{tikz}
\usepackage{tikz-layers}
\usetikzlibrary{arrows.meta}
\usetikzlibrary{positioning, calc}
\usepackage{forest}
\def\nodeminsize{8}
\def\nodedistance{30}
\tikzset{
    roundnode/.style={ellipse, draw=black!60, fill=black!2, thick, minimum size=\nodeminsize mm, node distance=\nodedistance mm, inner sep=1.75mm, font=\normalsize},
    rectanglenode/.style={rectangle, draw=black!60, fill=black!2, thick, minimum size=\nodeminsize mm, node distance=\nodedistance mm, inner sep=1.75mm, font=\normalsize},
    blanknode/.style={rectangle, draw=black!0, fill=black!0, thick, minimum size=\nodeminsize mm, node distance=\nodedistance mm, inner sep=1.75mm, font=\normalsize},
    edge from parent/.style={draw,->,black},
    line/.style={->},
    missionnode/.style={ellipse, draw=black!60, fill=black!5, thick, minimum size=8 mm, node distance=40 mm, inner sep=1.5mm, font=\normalsize, align=center}, 
}

\makeatletter
\def\@ssect#1#2#3#4#5#6{%
  \NR@gettitle{#6}
  \old@ssect{#1}{#2}{#3}{#4}{#5}{#6}
}
\makeatother

\begin{document}
\begin{frontmatter}
\title{Reactive Task Allocation for Balanced Servicing of Multiple Task Queues}

\author[First]{Niklas Dahlquist}
\author[First]{Akshit Saradagi} 
\author[First]{George Nikolakopoulos}

\address[First]{Luleå University of Technology, Luleå, SE-971 87 Sweden (e-mail: nikdah, akssar, geonik@ ltu.se).}
\begin{abstract}
In this article, we propose a reactive task allocation architecture for a multi-agent system for scenarios where the tasks arrive at random times and are grouped into multiple queues. Two stage tasks are considered where every task has a beginning, an intermediate and a final part, typical in pick-and-drop and inspect-and-report scenarios. A centralized auction-based task allocation system is proposed, where an auction system takes into consideration bids submitted by the agents for individual tasks, current length of the queues and the waiting times of the tasks in the queues to decide on a task allocation strategy. The costs associated with these considerations, along with the constraints of having unique mappings between tasks and agents and constraints on the maximum number of agents that can be assigned to a queue, results in a Linear Integer Program (LIP) that is solved using the SCIP solver. For the scenario where the queue lengths are penalized but not the waiting times, we demonstrate that the auction system allocates tasks in a manner that all the queue lengths become constant, which is termed balancing. For the scenarios where both the costs are considered, we qualitatively analyse the effect of the choice of the relative weights on the resulting task allocation and provide guidelines for the choice of the weights. We present simulation results that illustrate the balanced allocation of tasks and validate the analysis for the trade-off between the costs related to queue lengths and task waiting times.
\end{abstract}

\end{frontmatter}
\section{Introduction}
\subsection{Motivation}
The use of autonomous robots has seen a big increase in the recent years, showing promising results in areas such as search and rescue (\cite{9220149})
and warehouse management (\cite{9410352, 8901065}). Teams of collaborative autonomous agents are often used in such real-world large-scale scenarios. The complexity of deploying and coordinating multiple agents is a challenge, requiring both competent agents with communication capabilities, advanced sensors and on-board computers; and a sophisticated system for the management and coordination of the entire system. These challenges become even more demanding when operating in a dynamic environment where the overall mission or the current tasks that are to be completed are unknown or varying. In this scenario, the system in charge of coordinating a team of robots must be able to reactively adjust to possible changes, both related to the environment and to the fact that the available tasks might change with time. Another important aspect of deploying multiple agents in a space shared among the agents and with other entities, is that the possibility of collision between agents and with other entities is very high.

In scenarios such as truck management in mines (\cite{truck_management_mines}), computer networks (\cite{example_network_balancing}) and in fulfillment centers (\cite{multi_robot_e_commerce_}), the tasks to be executed are ordered in multiple in queues/picking stations where the tasks have to be assigned to agents in a manner that every queue is fairly serviced in an equalized manner, even in a dynamic setting. An optimal task allocator faces two main challenges in this scenario: balancing task allocation in servicing different queues and making sure that no task in a queue has to wait too long before being serviced.
\subsection{Background}
There are many approaches for coordinating teams of autonomous agents and many of these have been an active field of research in robotics recently (\cite{9029554, 9220149}). Many of the current state-of-the-art solutions consider a scenario where the tasks to be solved are static, or known in advance (\cite{6213575}). Some problems include vehicle routing and coalition formation (\cite{https://doi.org/10.48550/arxiv.2207.09650, https://doi.org/10.48550/arxiv.1705.10868}). The popular approaches to solving the problem of multi-agent task allocation are Optimization-based, deterministic approach (\cite{6225234}), market-based approach (\cite{9029554, 5980500}) and learning-based approach (\cite{9116987}).

The problem under consideration in this article concerns task allocation for managing task queues in a dynamic scenario where there is no a priori information of the tasks to be solved. Queue management is a well studied problem, especially in computer science. In network routers, there are several standard ways of handling dynamic inflow of data from different sources with different priorities. Some of the popular methods include: First-In First-Out (FIFO), Custom Queuing (CQ), Priority Queuing (PQ), Weighted Fair Queuing (WFQ) and Class Based Weighted Fair Queuing (CBWFQ) (\cite{8632161}).

The agents part of a multi-agent system must possess a wide array of behaviours to interact with other agents and the dynamic environment. Behavior trees, having their origin in computer games 
, are being increasingly used within the scope of robotics. Behaviour trees have the ability of generate modular and reconfigurable autonomy for robotic agents 
and are shown to generalize other well known decision architectures such as finite state machines and decision trees (\cite{BT_introduction}). 

In this article, we propose an architecture that unifies behavior trees for controlling individual agents with a centralized market-based auction system, for the multi-agent task allocation problem for balanced servicing of multiple task queues. The centralized task-allocator makes sure that the entire system optimally works as a team towards executing the available tasks, but without needing to know or be involved in the finer details of the agent-level task execution. The behavior trees allow individual agents to act independently and be responsible for local coordination (such as local collision avoidance) without needing to know the bigger picture of the overall mission.

\subsection{Contributions}
The following are the main contributions of this article.
\begin{enumerate}
\item In this article, we propose a novel market-inspired reactive approach to solve the multi-agent pick up and delivery task allocation problem, for the scenario where the tasks arrive and are ordered into multiple queues/pickup stations. The approach considers the lengths of the queues and waiting times of the items at the queues, along with the path costs of the agents to balances the load on multiple pickup stations.
\item Different choices for the relative weights between path costs, queue lengths and waiting times leads to different evolutions of the queues. This is due to the coupling and trade-offs involved in the costs being considered. We perform and present a qualitative analysis of the effect of the relative weight on the possible behaviors and the present guidelines for the choice of different parameters.
\item The behavior trees framework is integrated into the task allocation architecture, for managing individual agents. The current behavior tree state is used to indicate the availability of an agent for a given task and for calculating the cost of completing a task.
\item We present extensive simulation based evaluation of the proposed approach through illustrative scenarios and verify our analysis of the effects of the design parameters on the resulting behavior of the system.
\end{enumerate}
\section{Problem Formulation}
The set of picking stations or queue locations is denoted by \(\mathfrak{S} = \{S_1, \dots, S_{n_p}\}\), where \(n_p \in \mathbb{N}\) is the total number of picking stations and \(S_i\), $i\in\{1, \ldots, n_p\}$, denote the individual picking stations. Each \(S_i\) is associated with a set of currently available pick up and deliver tasks \(\mathfrak{Q}_i\), $i\in\{1, \ldots, n_p\}$. The set of all available tasks from all the queues is then formed as \(\mathcal{T}  = \mathfrak{Q}_1 \cup \dots \cup \mathfrak{Q}_{n_p} = \{T_1, \dots T_{n_t}\}\) where \(n_t \in \mathbb{N}\) is the total number of available tasks and $T_j$, $j\in\{1, \ldots, n_t\}$, denotes the individual tasks. A team of multiple ground agents \(\mathfrak{R} = \{R_1, \dots, R_{n_a}\}\), where \(n_a \in \mathbb{N}\) is the total number of agents, are available for finishing the tasks. Each task in the set of tasks \(\mathcal{T}\) consists of a target pickup location and a target drop-off location. We consider the homogeneous task and homogeneous agent scenario, where every agent \(R_k\) is capable of completing any given task \(T_j\). A given task is said to be completed when an agent moves from its current position to a desired pickup location, retrieves an item, moves to the target drop-off location and leaves the item at the specified position. The action of picking up an item is considered irreversible, in the sense that once an agent have picked up an item, it must drop it off before being able to pick up another item. 

Additionally, the following assumptions are made for the precise definition of the scenario being considered.

\emph{Assumption 1:} Each agent is able to complete the assigned pick up and deliver tasks without failing or crashing. Abandoning a task before completion is not considered, except when the auction system actively reassigns an agent to a different task, before the pick up operation. 

\emph{Assumption 2:} There exists a reliable two-way communication between every agent and the centralized task allocator. This communication link is used by the agents to send bids to the task allocator and by the task allocator to send task assignments to agents.

\emph{Assumption 3:} Every agent has the computational resources and a map of the environment, to estimate the cost for completing each task, given the pick up and drop-off locations.

\emph{Assumption 4:} Every station keeps track of the number of waiting tasks in its queue, i.e., the queue length. This is known by the task allocator at every auctioning phase.

%
%
\begin{figure}[t]
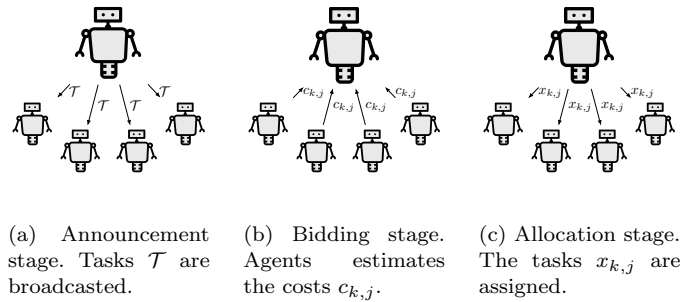

    \centering
    \begin{subfigure}[t]{0.3\columnwidth}
        \centering
        \input{figures/tex/auction_fig_1}
        \caption{Announcement stage. Tasks \(\mathcal{T}\) are broadcasted.}
    \end{subfigure}
    \hfill
    \begin{subfigure}[t]{0.3\columnwidth}
        \centering
        \input{figures/tex/auction_fig_2}
        \caption{Bidding stage. Agents estimates the costs \(c_{k, j}\).}
    \end{subfigure}
    \hfill
    \begin{subfigure}[t]{0.3\columnwidth}
        \centering
        \input{figures/tex/auction_fig_3}
        \caption{Allocation stage. The tasks \(x_{k, j}\) are assigned.} %
    \end{subfigure}
    \caption{An illustration of the different stages of the auction system.}
    \label{fig:auction_summary}
\end{figure}
\subsection{Market-inspired Auction System}
The market-inspired auction system for optimal task allocation is illustrated in Fig. \ref{fig:auction_summary} and the three stages can be summarized as.
1) Announcement stage: The available tasks in the set $\mathcal{T}$ are announced to the agents.
2) Bidding stage: costs $c_{k, j}$, the cost of agent $k$ performing task $j$, are calculated by the agents for the available tasks. The costs $c_{k, j}$ are then sent as bids to the centralized task allocator.
4) Task allocation stage: The central unit collects the bids from the agents and the status of the tasks and queues from the pickup stations, optimizes and announces the allocations $x_{k, j}$. The variable $x_{k, j}=1$ implies agent $R_k$ has been assigned to task $T_j$.

\emph{Problem Definition:}
The definitions and assumptions described so far define the setting of the problem. In this setting the problem definition is stated as follows.

The entry of new tasks into the auctioning system is not known a priori and new tasks enter into the system in a random manner and are assigned a picking station \(S_i\) and are put into the collection of available tasks \(\mathcal{T}\). 
For deciding the optimal task assignments, the centralized auction system minimizes the total cost taking into account the bids from the agents, the cost of the current lengths of the queues and the cost associated with the waiting time of every task. This is done reactively every time a new task enters into one of the queues. More precisely, the auction system minimizes the following cost
\begin{equation}
    \min_{x_{k, j} \in \{0, 1\}} \sum_{k,j\in E}(\sum_{\{k \mid R_k \in \mathfrak{R}\}} c_{k,j} + \sum_{\{i \mid i \in \mathfrak{S}\}} q_{i} + \sum_{\{j \mid j \in \mathcal{T}\}} \tau_{j}) \cdot x_{k, j}
    \label{cost_function}
\end{equation}
where $E=\{1,\ldots,n_a\}\times\{1,\ldots,n_t\}$, \(x_{k, j}\) is an assignment variable representing if agent \(R_k\) is allocated to task \(T_j\), \(c_{j,k}\) represents the cost of allocating agent \(R_k\) to task \(T_j\). If $T_j$ is part of the queue at station $S_i$, that is $T_j\in\mathfrak{Q}_i$, then 
\(q_{i}\) represents the penalty of the station \(S_{i}\) that is proportional to its current queue length and \(\tau_{j}\) represents the cost associated with the time that the task \(T_j\) has been waiting to be completed in station $S_i$.
We consider the following constraints for the optimization problem \eqref{cost_function}.
\begin{equation}
    \begin{aligned}
        \sum_{\{k \mid R_k \in \mathfrak{R}\}} x_{k,j} &\leq 1, \text{ }  \forall j \in \{1, \ldots, n_t\} \\
        \sum_{\{j \mid T_j \in \mathcal{T}\}} x_{k,j} &\leq 1, \text{ }  \forall k \in \{1, \ldots, n_a\} \\
        \sum_{\{j \mid T_j \in \mathfrak{Q}_i\}} x_{k,j} &\leq m_i, \text{ } \forall k \in \{1, \ldots, n_a\} \;\&\; \forall i \in \{1, \ldots, n_p\}
    \end{aligned}
    \label{constraints}
\end{equation}
The first two constraints make sure that only one agent is assigned to a task and no task is assigned to multiple agents. The third constraint ensures that not more than \(m_i\) agents are assigned to tasks from \(\mathfrak{Q}_i\) belonging to a service station $S_i$. The function penalizing the queue length, \(q_i\), and the waiting time, \(\tau_j\) are chosen to be directly proportional to the queue lengths and waiting times respectively,
\begin{equation}
    q_i = q \cdot | \mathfrak{Q} |
\end{equation}
and 
\begin{equation}
    \tau_j = \tau \cdot | \mathcal{T} |
\end{equation}
where \(q\) and \(\tau\) are constants. The $|\cdot|$ operation over a set yields the number of elements in the set.

\section{Methodology}

\subsubsection{Optimization Problem :} The problem of deciding what task to allocate to which agent in a way that minimizes the total cost \eqref{cost_function} subject to constraints \eqref{constraints} and assigns the maximum number of tasks is solved by creating an undirected graph \(G = (N, E)\) where \(N\) represents nodes and \(E\) represents edges between the nodes. The nodes \(N\) consists of both the agents \(R_k\) and all the available tasks \(T_j\); the edges \((m, l) \in E\) represents the cost of a potential match between node \(m\) and \(l\). The edges between agents \(R_k\) and tasks \(T_j\) represents the cost
\begin{equation}
    C_{k, j} = c_{k, j} + q_i + \tau_j.
\end{equation}
To optimize the cost and at the same time assign the maximum number of tasks, the costs \(C_{k,j}\) are converted to profits
\begin{equation}
    \rho_{k,j} = \text{profit of assigning agent \(R_k\) to task \(T_j\)}
\end{equation}
in a way that preserves the inverse order of the costs. The assignment variable
\begin{equation}
    x_{k,j} = 
    \begin{cases}
        1,  \text{ if node \(k\) is matched to node \(j\)}\\
        0, \text{ otherwise}
    \end{cases}, (k,j) \in E
\end{equation}
is introduced as an binary decision variable that describes if agent \(R_k\) is assigned to task \(T_j\). The objective function can then be compactly written as
\begin{equation}
    \max_{x_{k, j} \in \{0, 1\}, (k, j) \in E} \sum_{(k,j) \in E} \rho_{k,j} \cdot x_{i,j}.
    \label{eq:auction_optimization}
\end{equation}
\subsection{Task Execution by Agents}

\subsubsection{Behavior Trees:}
A behavior tree is a model for how to dynamically switch between a number of behaviors to create a reactive system, it is constructed as a number of nodes connected as a directed acyclic graph, where the internal nodes are called \emph{control nodes} and the leaf nodes \emph{execution nodes}. The behavior tree is executed by "ticking" the root node, the "tick" is then passed down the tree by the control nodes until an execution node receives it. The types of nodes and the execution details are described in \cite{BT_introduction}.

\subsubsection{Path Planning:}
In the context of a pick up and deliver task, the fundamental cost to be considered in the task allocation framework is the distance of the shortest path for completing the task. Additionally, one could consider the shortest safe path as well, where the planning problem includes if the robot can safely travel through the path, for a more realistic evaluation of how the agent should reach that goal safely. The grid-search algorithm DSP~\cite{karlsson2021d}, based on D\(^*\) Lite, is a path planner that incorporates a risk-layer to avoid entering high-risk areas (defined by being close to an occupied cell) if not absolutely necessary to reach the goal. We can say that DSP plans a path $P$ from robot position $\hat{p}$ to the task goal $p_g$ such that  $\sum{\forall \Gamma_{\hat{p} \to p_g} \in \vec{P}_{\hat{p} \to p_g}}$ is minimized, where the cost $\Gamma = \Gamma_\mathrm{dist} + \Gamma_\mathrm{risk}$. 
In the following scenarios, DSP is used both for generating safe paths and for estimating the distance to various positions.

\subsubsection{Cost for Executing Tasks:}
To estimate the costs for completing individual tasks, each agent defines a cost functions that estimates the cost of executing that task.
The cost function depends on the state of the agent, the properties of the agent and the specific type of task. 
It is not required that a cost function is defined for every available task, since the agent needs only to define a cost function for the tasks that it can execute. 
A general cost function, describing the cost for agent \(R_k\) to perform task \(T_j\) is defined as
\begin{equation}
    c_{k, j} = f_{k, j} (\text{task}, \text{ state}) \cdot k_\text{bt}
\end{equation}
where \(k_\text{bt}\) is a factor that depends on the state of the current task, derived from the state of the running behavior tree.

\subsection{Details and Specifications of the Simulation Setup}

\subsubsection{Task Considered:}
The task that is considered in this work is the "pick up and deliver" task. 
More specifically, a task consists of a pickup location and a drop-off location and the agent is required to first move to the pickup location and then move to the drop-off location. Additionally, while moving towards the drop-off location, the agent can not switch to another task until the current one is completed.

\subsubsection{Behavior Trees Used:}
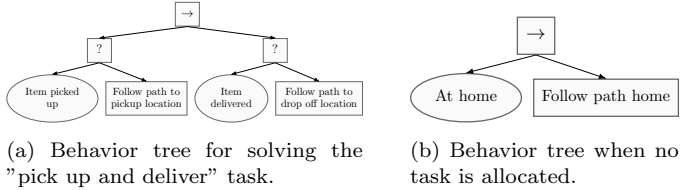
\begin{figure}
    \centering
        \begin{subfigure}[t]{0.53\columnwidth}
        \centering
    \begin{adjustbox}{max width = \columnwidth}
        \begin{forest}
        for tree={align=center, edge={-{Latex[length=1.5mm]}, draw}, parent anchor=south,child anchor=north}
        [\large \(\rightarrow\), rectanglenode
            [\large ?, rectanglenode
                [Item picked\\up, roundnode]
                [Follow path to\\pickup location, rectanglenode]
            ]
            [\large ?, rectanglenode
                [Item\\delivered, roundnode]
                [Follow path to\\drop off location, rectanglenode]
            ]
        ]
        \end{forest}
    \end{adjustbox}
        \caption{Behavior tree for solving the "pick up and deliver" task.}
        \label{fig:bt_all:pickup_deliver}
    \end{subfigure}
    \hfill
    \begin{subfigure}[t]{0.40\columnwidth}
        \centering
    \begin{adjustbox}{max width = \columnwidth}
        \begin{forest}
        for tree={align=center, edge={-{Latex[length=1.5mm]}, draw}, parent anchor=south,child anchor=north}
        [\large \(\rightarrow\), rectanglenode
            [At home, roundnode]
            [Follow path home, rectanglenode]
        ]
        \end{forest}
    \end{adjustbox}
        \caption{Behavior tree when no task is allocated.}
        \label{fig:bt_all:no_task}
    \end{subfigure}

    \caption{The behavior trees used to in the evaluation scenarios.}
    \label{fig:bt_all}
\end{figure}

The agents execute a task specific behavior tree when they are allocated to a task and another behavior tree when they are not allocated to a task. The behavior tree specific to the task "pick up and deliver" will make sure that the agent finishes the task, the behavior tree associated with not being allocated to a task will result in that the agent moves back to its starting position. The behavior tree for executing the task "pick up and deliver" is shown in Fig. \ref{fig:bt_all:pickup_deliver} and the behavior tree that is executed otherwise is shown in Fig. \ref{fig:bt_all:no_task}.

%
%
\subsubsection{Specific Cost Functions:}
The cost functions used for the task "pick up and deliver" is chosen as
\begin{equation}
    f_{k, j} (\text{task, state}) = (\Gamma_\text{dist, 1} + \Gamma_\text{dist, 2}) \cdot k_\text{bt}
\end{equation}
where \(\Gamma_\text{dist, 1}\) is the distance of the planned path from the current position to the pickup location, \(\Gamma_\text{dist, 2}\) is the distance from the pickup location to the drop-off location. I.e. the total distance required to move for solving the tasks. The factor \(k_\text{bt}\) is based on if the agent has already picked up an object, if that is the case that agent will reduce the cost for its current task to zero, by setting \(k_\text{bt} = 0\), to guarantee that the current task is not redistributed to another agent. Otherwise, \(k_\text{bt} = 1\).

\subsubsection{Simulation Environment} 
To evaluate the proposed architecture, Gazebo\footnote{https://gazebosim.org} is used. Gazebo is an open source physics simulator widely used within the robotics community and is tightly integrated with Robot Operaing System (ROS)\footnote{https://www.ros.org/}. This enables an realistic simulation environment to evaluate the performance and gives a good indication how the performance would translate to a real world scenario.

\subsubsection{Specific Parameters for the Auction System}

To evaluate the specific problem that is considered the parameters for the task allocation optimization where chosen as
\begin{equation}
    q_i = 10000 \cdot |\mathfrak{Q}_i|,
\end{equation}
\begin{equation}
    \tau_j = 100 \cdot |\mathcal{T}|,
\end{equation}
and 
\begin{equation}
    m_i = 1
\end{equation}
to give a balance between penalizing long queues and long waiting times for individual tasks, Since in the evaluation scenarios that will be considered the queue lengths are approximately three orders of magnitude larger compared to the waiting time for individual tasks.

\section{Simulation Results}
\begin{figure}
    \centering
    \includegraphics[width = 0.6\columnwidth]{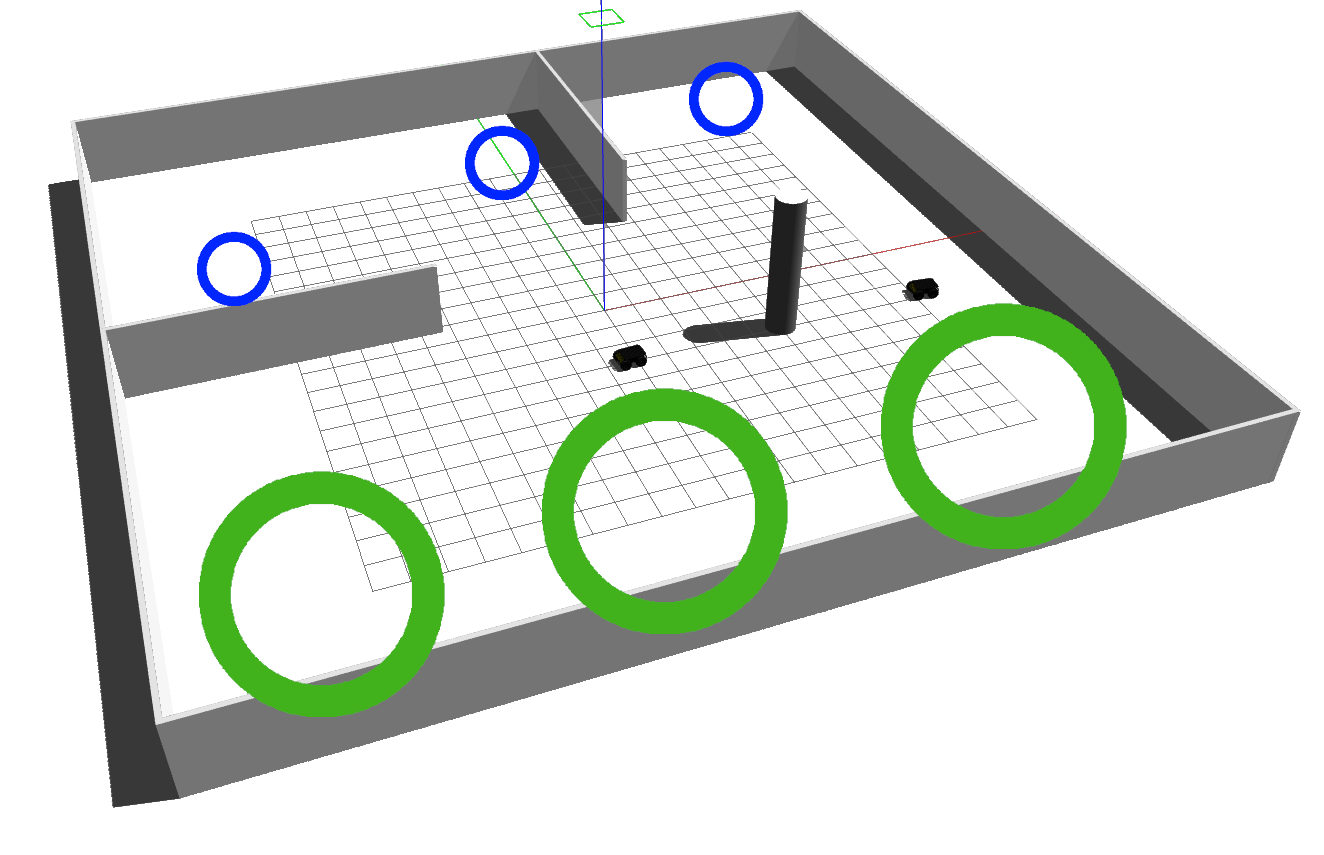}
    \caption{Initital configuration of the environment used in the evaluation scenarios, showing where the mobile agents starts. The pickup locations are marked by green circles and the drop-off locations by blue circles.}
    \label{fig:environmen_initial_gazebo}
\end{figure}

The previously explained framework is evaluated using a number of illustrative scenarios. The different scenarios aim to demonstrate the properties of the framework. In all scenarios there are three different pickup stations and three different drop-off locations, shared by all stations. The first scenario considers the case where a fixed number of tasks, with a random drop-off location associated with each task, are added to each station at start. The second scenario considers the same three pickup stations but with unequal number of tasks added at start. In the third, fourth and fifth scenarios tasks are dynamically inserted to the different queues at random times, these scenarios also compares the effects of omitting the penalty related to the time individual tasks have been waiting to be serviced in the queues \(\tau_j\) and omitting the penalty related to queue lengths \(q_i\) from the optimization problem. Fig. \ref{fig:environmen_initial_gazebo} shows the environment, and the initial configuration, related to the evaluation scenarios. 

All scenarios are evaluated in a simulation environment using multiple Huskys\footnote{https://clearpathrobotics.com/husky-unmanned-ground-vehicle-robot/} as mobile agents and a static map is used by all agents for path planning and cost estimation. The experiments have been performed on a laptop with an AMD Ryzen 7 PRO 5850U, 32 GB RAM and Ubuntu 20.04 as the operating system. The proposed framework has been implemented using C++ programming within ROS. The behavior trees are created and executed using the framework BehaviorTree.CPP\footnote{https://github.com/BehaviorTree/BehaviorTree.CPP} and the optimization problem \eqref{eq:auction_optimization} is
solved using SCIP \cite{GamrathEtal2020OO}.

\subsection{Scenario 1}
\begin{figure}
    \centering
    \includegraphics[width = 0.72\columnwidth]{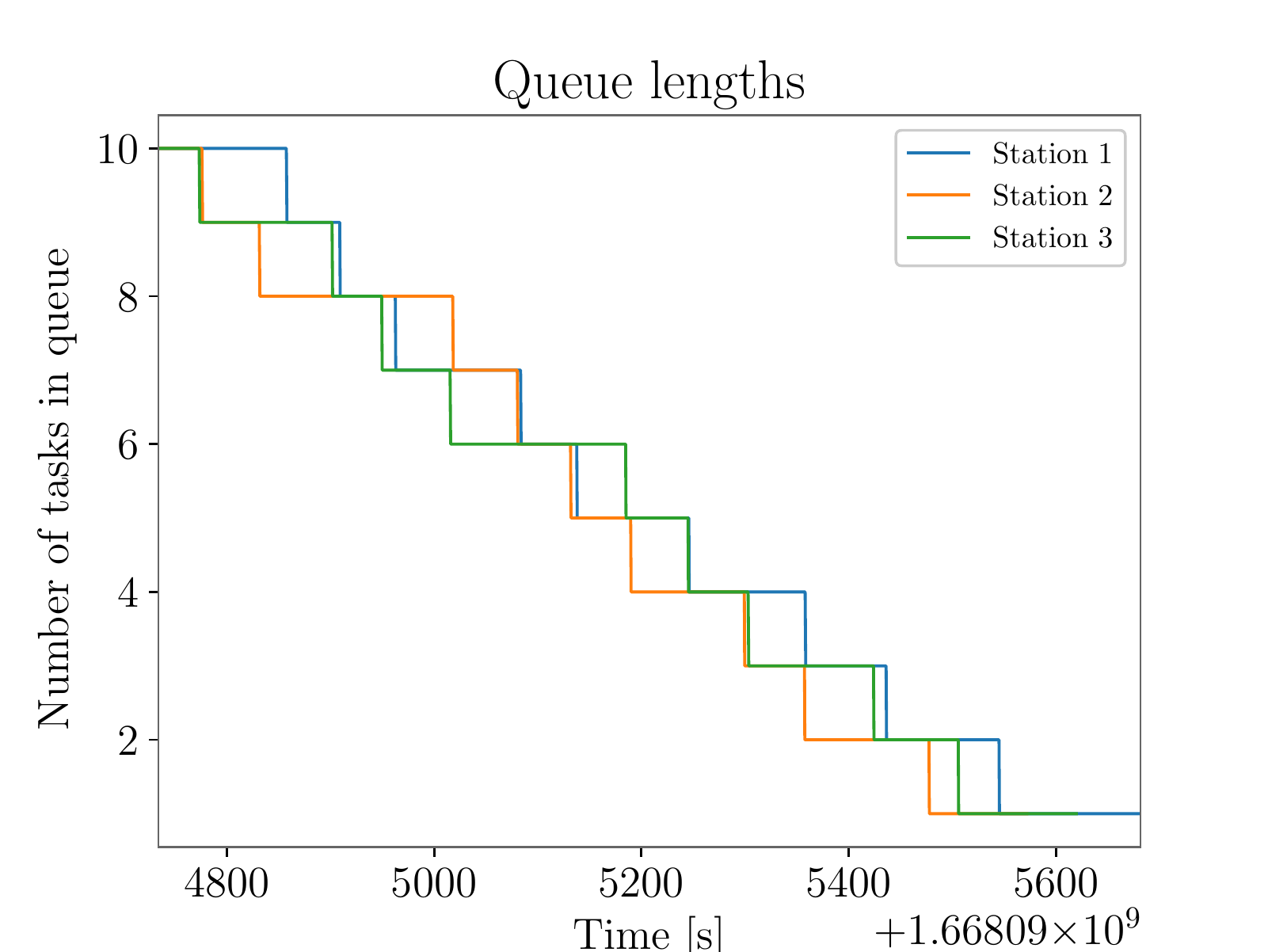}
    \caption{The evolution of the lengths of the queues during evaluation of scenario 1.}
    \label{fig:scenario_1_queue_lengths}
\end{figure}

Here, three stations and two mobile agents are used and 10 tasks are added to each station at start. The resulting lengths of each queue as all the tasks are being completed are shown in Fig. \ref{fig:scenario_1_queue_lengths}. As expected, the three queues are reduced to zero in a balanced manner with at most one task difference.

\subsection{Scenario 2}
\begin{figure}
    \centering
    \includegraphics[width = 0.72\columnwidth]{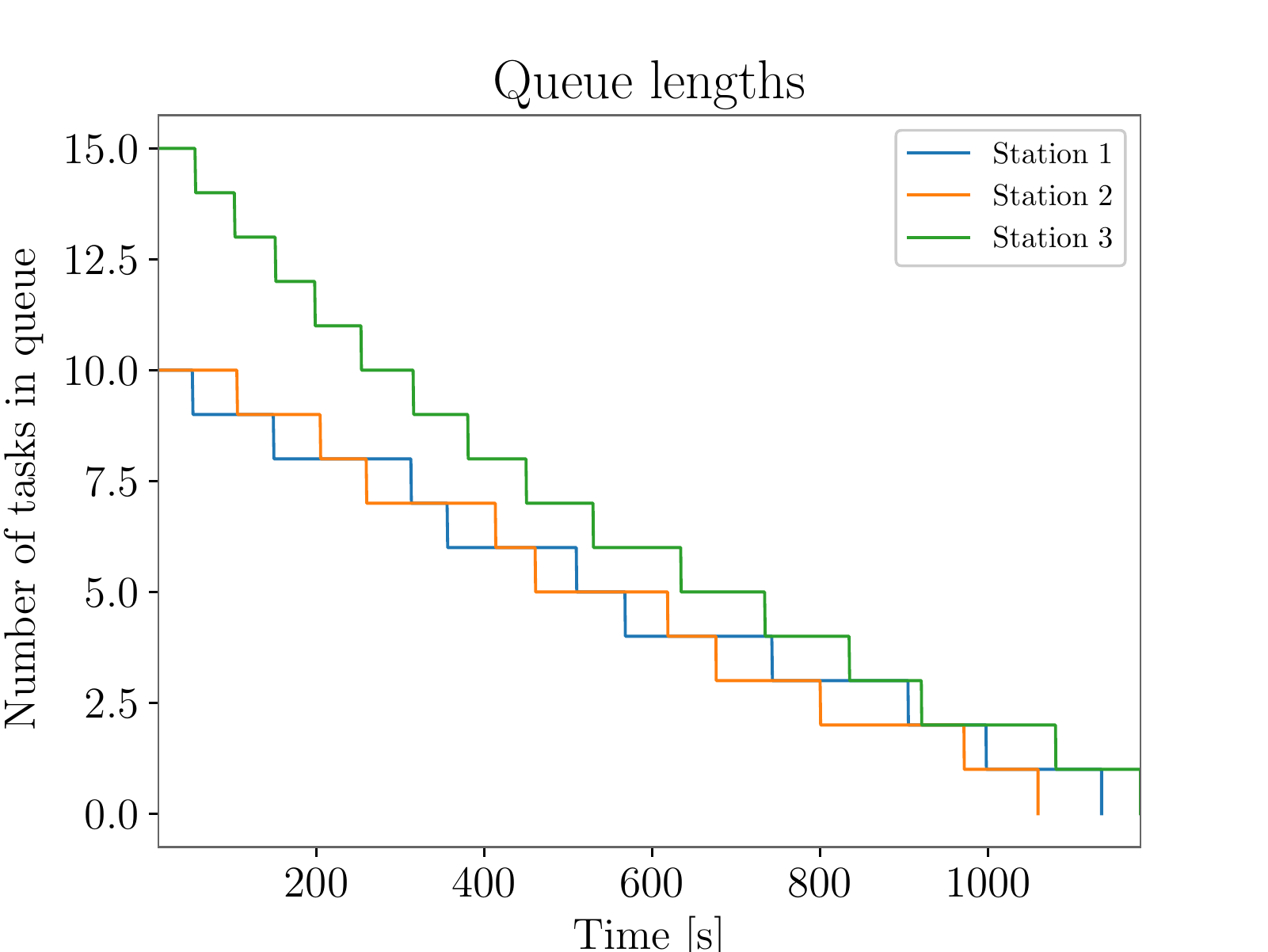}
    \caption{The evolution of the lengths of the queues during evaluation of scenario 2.}
    \label{fig:scenario_2_queue_lengths}
\end{figure}

In the second scenario three stations and two mobile agents are used and 10, 10, 15 tasks are added to each station, respectively, at start. The resulting lengths of each queue as all the tasks are being completed are shown in Fig. \ref{fig:scenario_2_queue_lengths}. It shows that the, in the beginning, unbalanced queues are converging to zero in a way that all tasks will be completed at roughly the same time.

\subsection{Scenario 3}
\begin{figure}[h]
    \centering
    \begin{subfigure}[t]{0.72\columnwidth}
        \centering
        \includegraphics[width = \columnwidth]{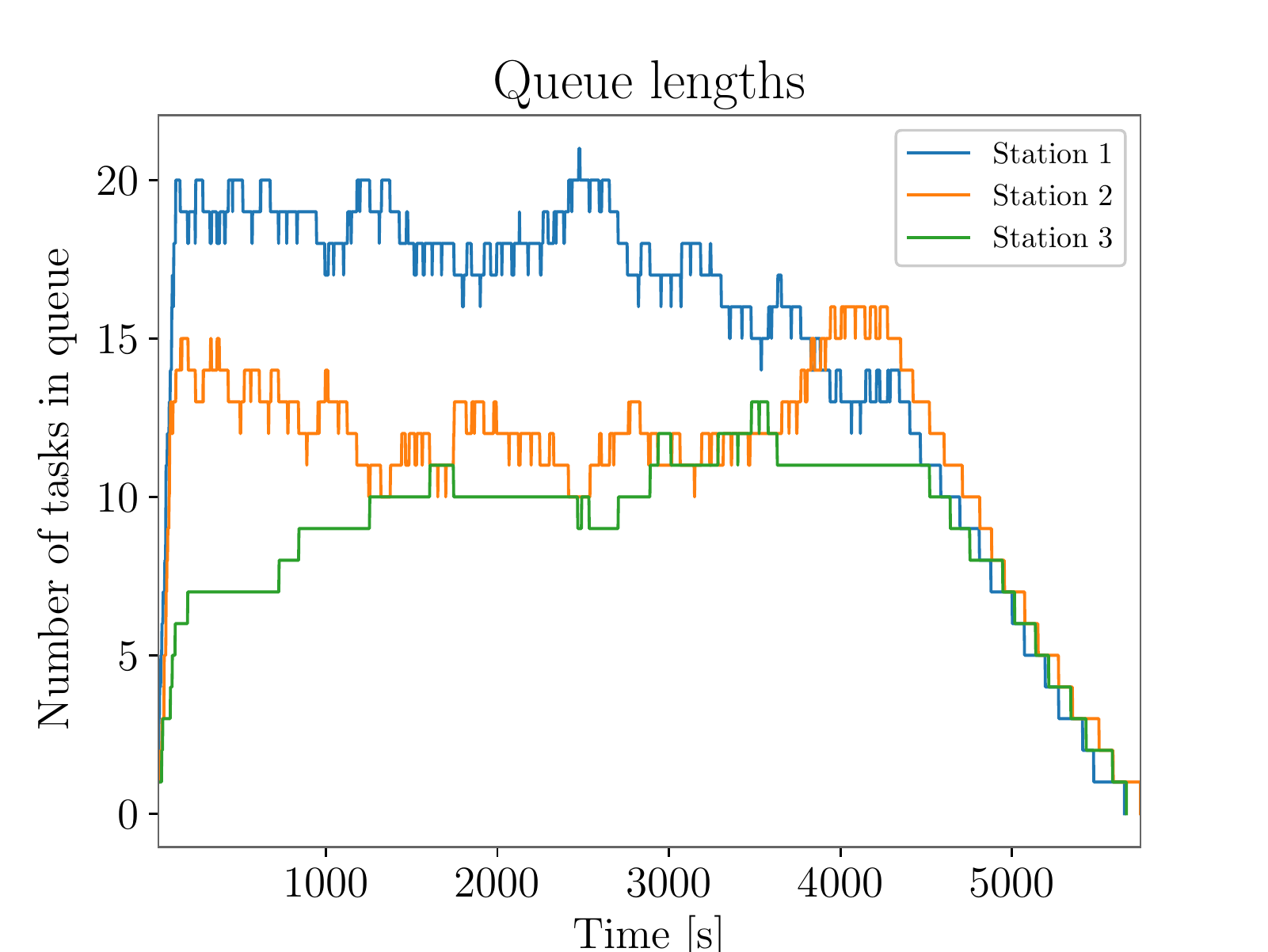}
        \caption{}
        \label{fig:scenario_3_all_figs:queue_length}
    \end{subfigure}
    \vfill
    \begin{subfigure}[t]{0.45\columnwidth}
        \centering
        \includegraphics[width = \columnwidth]{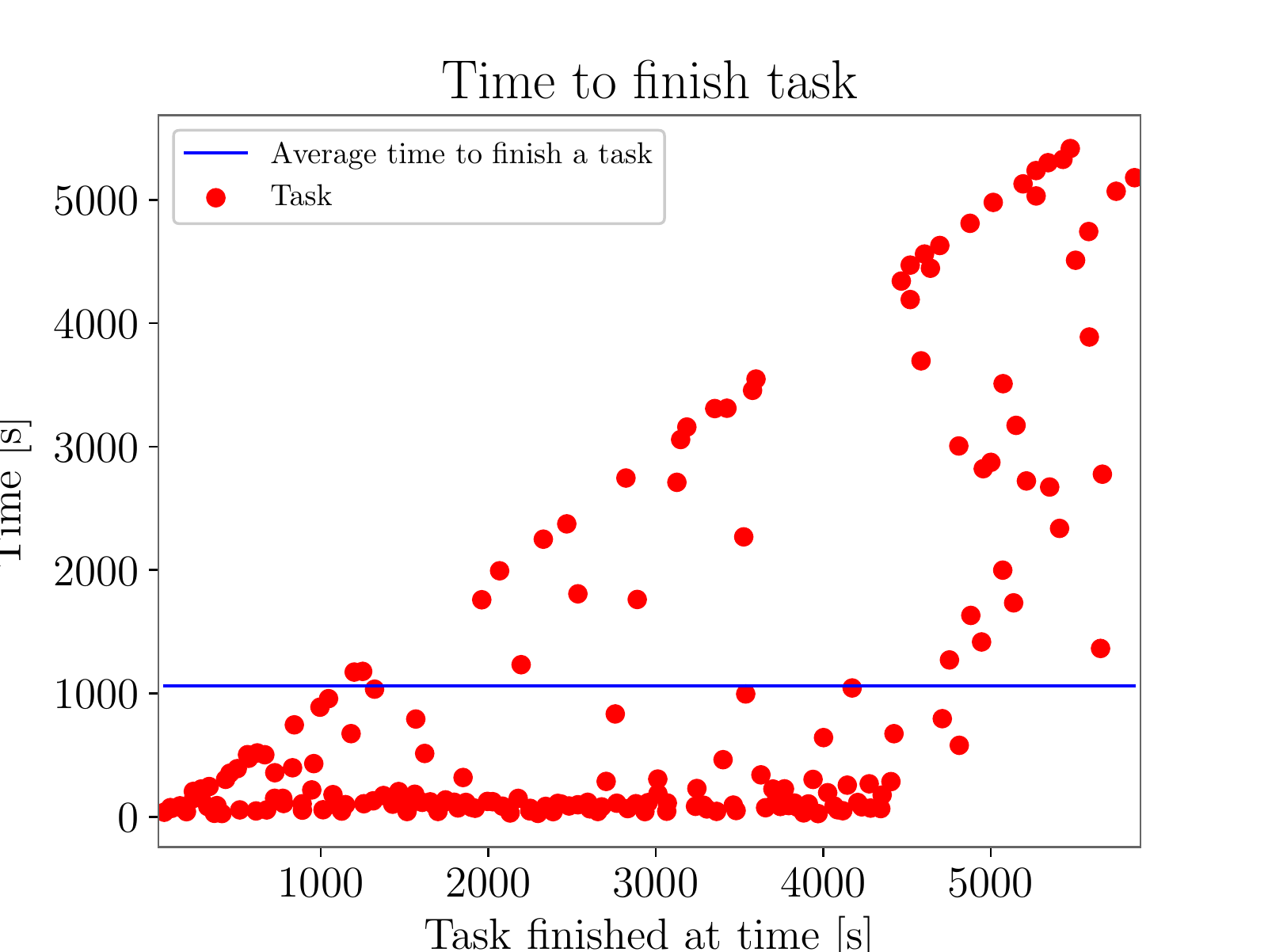}
        \caption{}
        \label{fig:scenario_3_all_figs:time_to_finish}
    \end{subfigure}
    \hfill
    \begin{subfigure}[t]{0.45\columnwidth}
        \centering
        \includegraphics[width = \columnwidth]{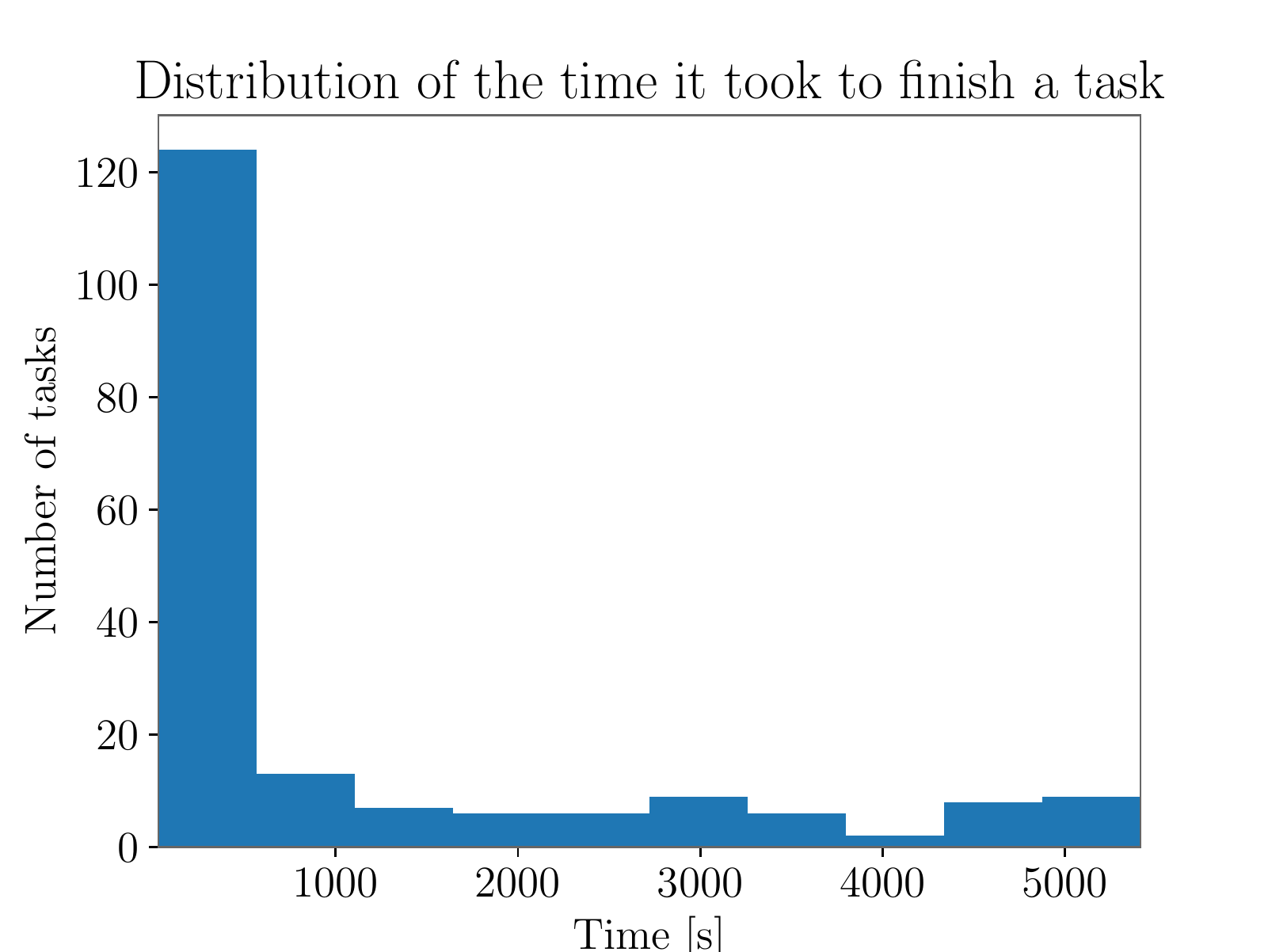}
        \caption{}
        \label{fig:scenario_3_all_figs:distribution}
    \end{subfigure}
    \caption{Summary of the results related to scenario 3. (a) The evolution of queue lengths during the evaluation scenario, (b) The times taken to complete the tasks after they have been added to the pool of available tasks and (c) A distribution of the times taken to complete the available tasks.}
    \label{fig:scenario_3_all_figs}
\end{figure}

In this scenario, three stations start with empty queues and two mobile agents are available for executing tasks. Every second there is a probability of \(5\%\), \(15\%\) and \(15\%\) for each, individual, queue to add a task. There is also a global maximum of the total number of tasks that can be available of \(40\) tasks, this means that if the maximum number of total tasks is reached, no more tasks can be added by any station. In this scenario the parameter \(\tau\) is adjusted to
\begin{equation}
    \tau = 0
\end{equation}
to demonstrate how the system behaves when not considering how long individual tasks have been available. The results are shown in Fig \ref{fig:scenario_3_all_figs} and it can be seen that all queues, after some time, are hovering at approximately the same length, however, the waiting time for some of the tasks are becoming very big relative to the average waiting time.

\subsection{Scenario 4}
\begin{figure}
    \centering
    \begin{subfigure}[t]{0.72\columnwidth}
        \centering
        \includegraphics[width = \columnwidth]{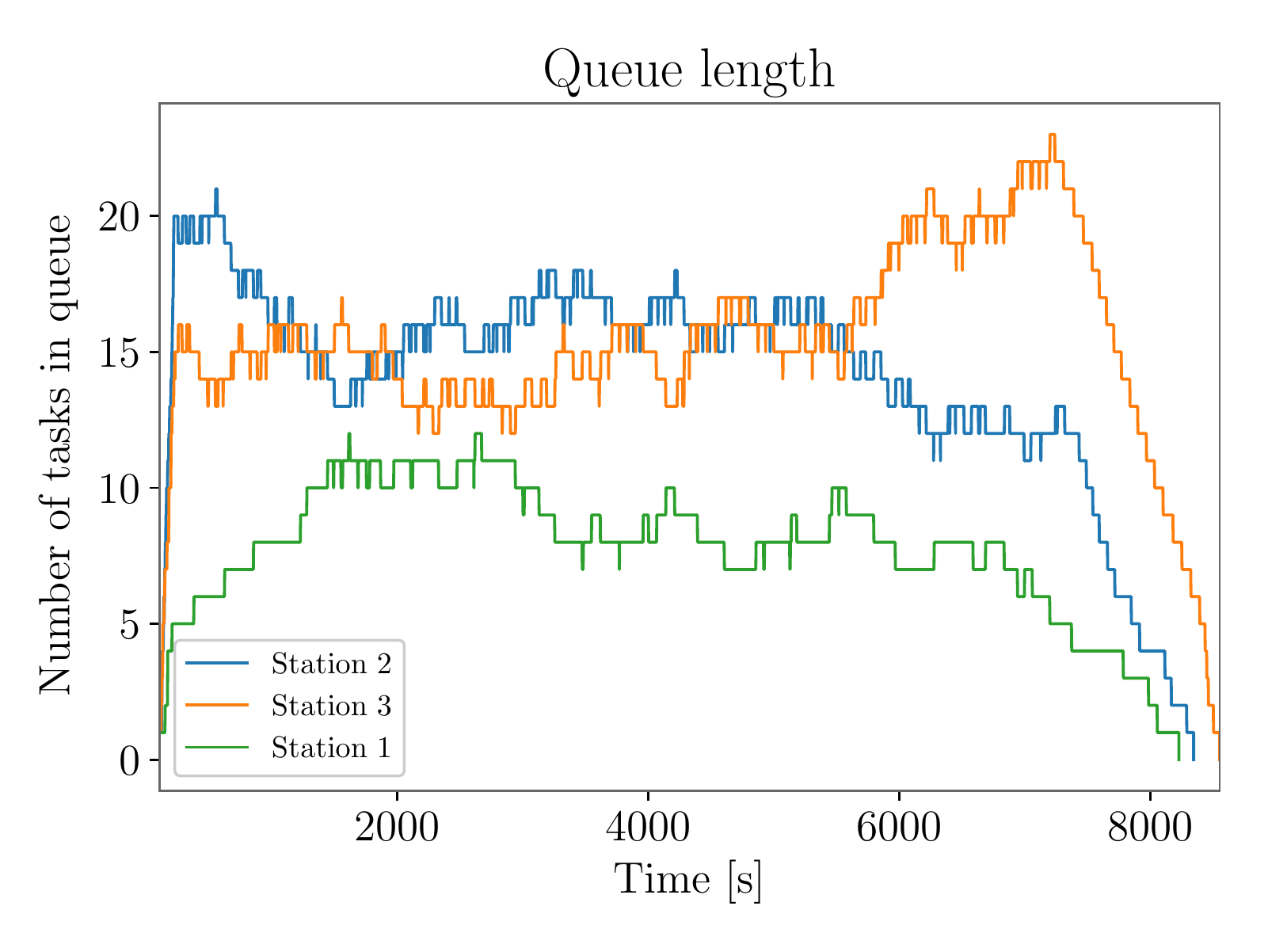}
        \caption{}
        \label{fig:scenario_4_all_figs:queue_length}
    \end{subfigure}
    \vfill
    \begin{subfigure}[t]{0.45\columnwidth}
        \centering
        \includegraphics[width = \columnwidth]{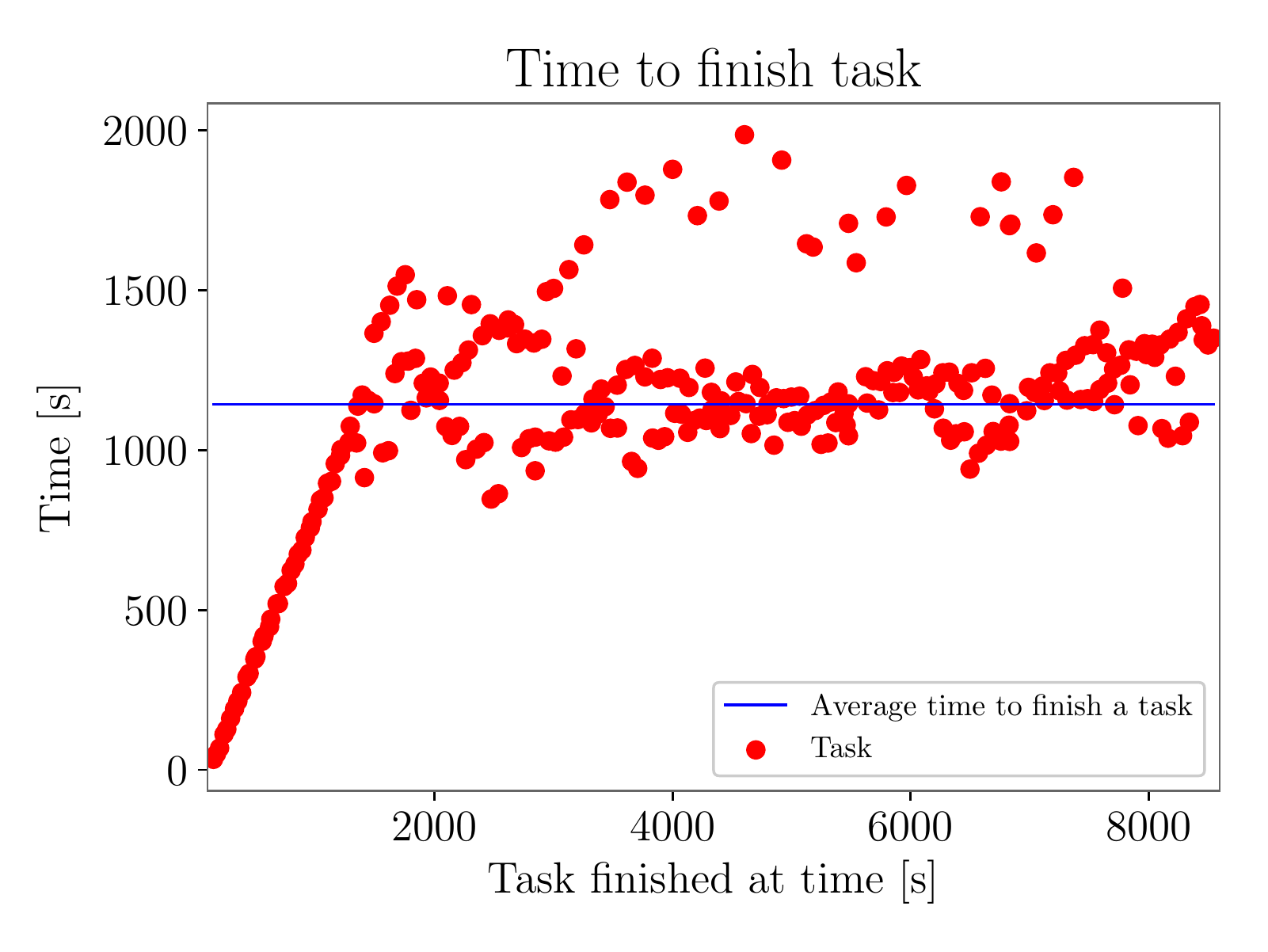}
        \caption{}
        \label{fig:scenario_4_all_figs:time_to_finish}
    \end{subfigure}
    \hfill
    \begin{subfigure}[t]{0.45\columnwidth}
        \centering
        \includegraphics[width = \columnwidth]{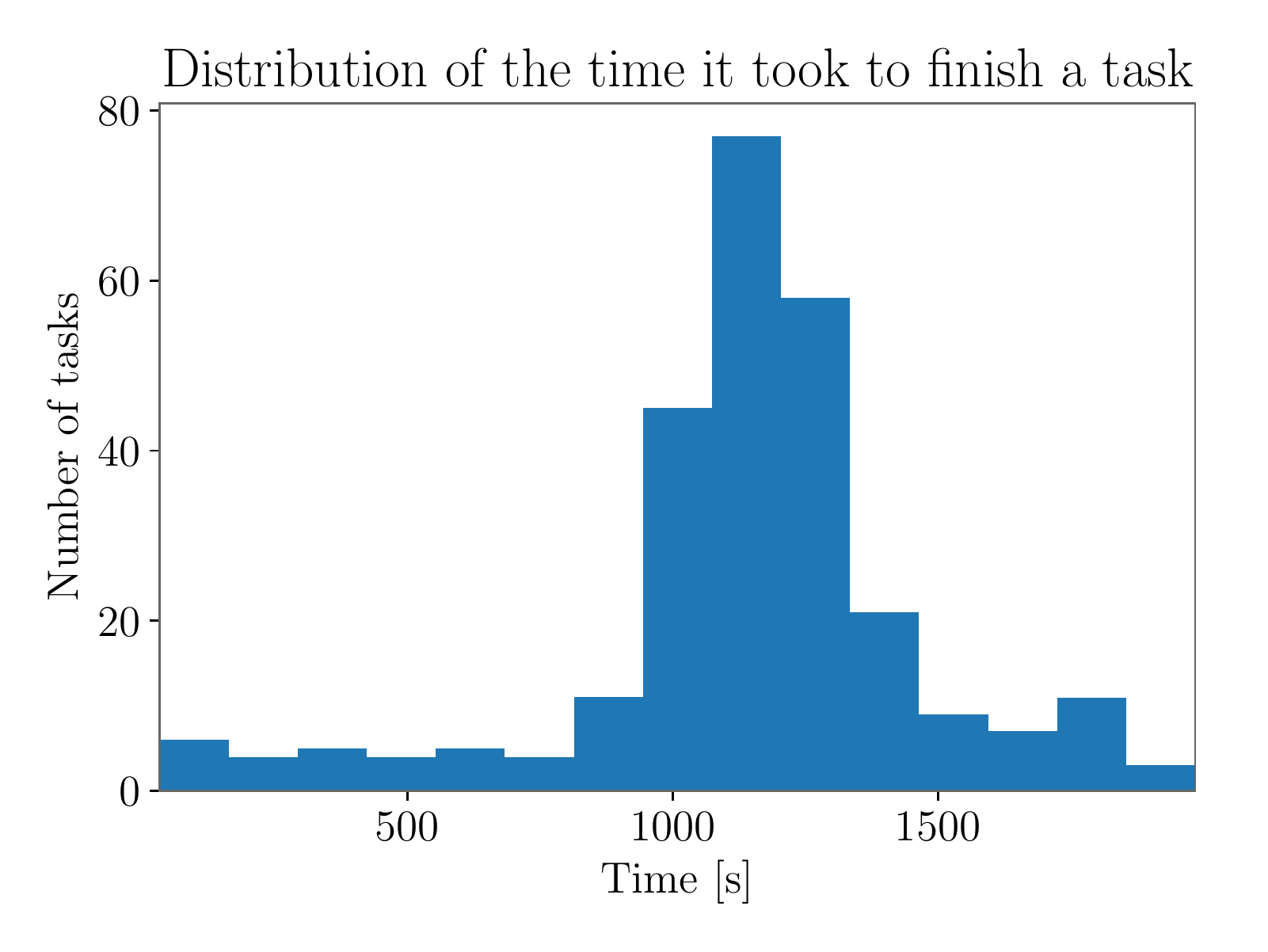}
        \caption{}
        \label{fig:scenario_4_all_figs:distribution}
    \end{subfigure}
    \caption{Summary of the results related to scenario 4. (a) The evolution of queue lengths during the evaluation scenario, (b) The time taken to complete the tasks after they have been added to the pool of available tasks and (c) A distribution of the times taken to complete the available tasks.}
    \label{fig:scenario_4_all_figs}
\end{figure}

In this scenario three stations start with empty queues and two mobile agents are available for executing tasks. Every second there is a probability of \(5\%\), \(15\%\) and \(15\%\) for each, individual, queue to add a task. There is also a global maximum of the total number of tasks that can be available of \(40\) tasks, this means that if the maximum number of total tasks is reached, no more tasks can be added by any station. This scenario setup is identical to scenario 3, except that we know include the penalty \(\tau_j\) related to how long every individual task have been in a queue. The results of this can be seen in Fig \ref{fig:scenario_4_all_figs}.

\subsection{Scenario 5}

\begin{figure}
    \centering
    \begin{subfigure}[t]{0.72\columnwidth}
        \centering
        \includegraphics[width = \columnwidth]{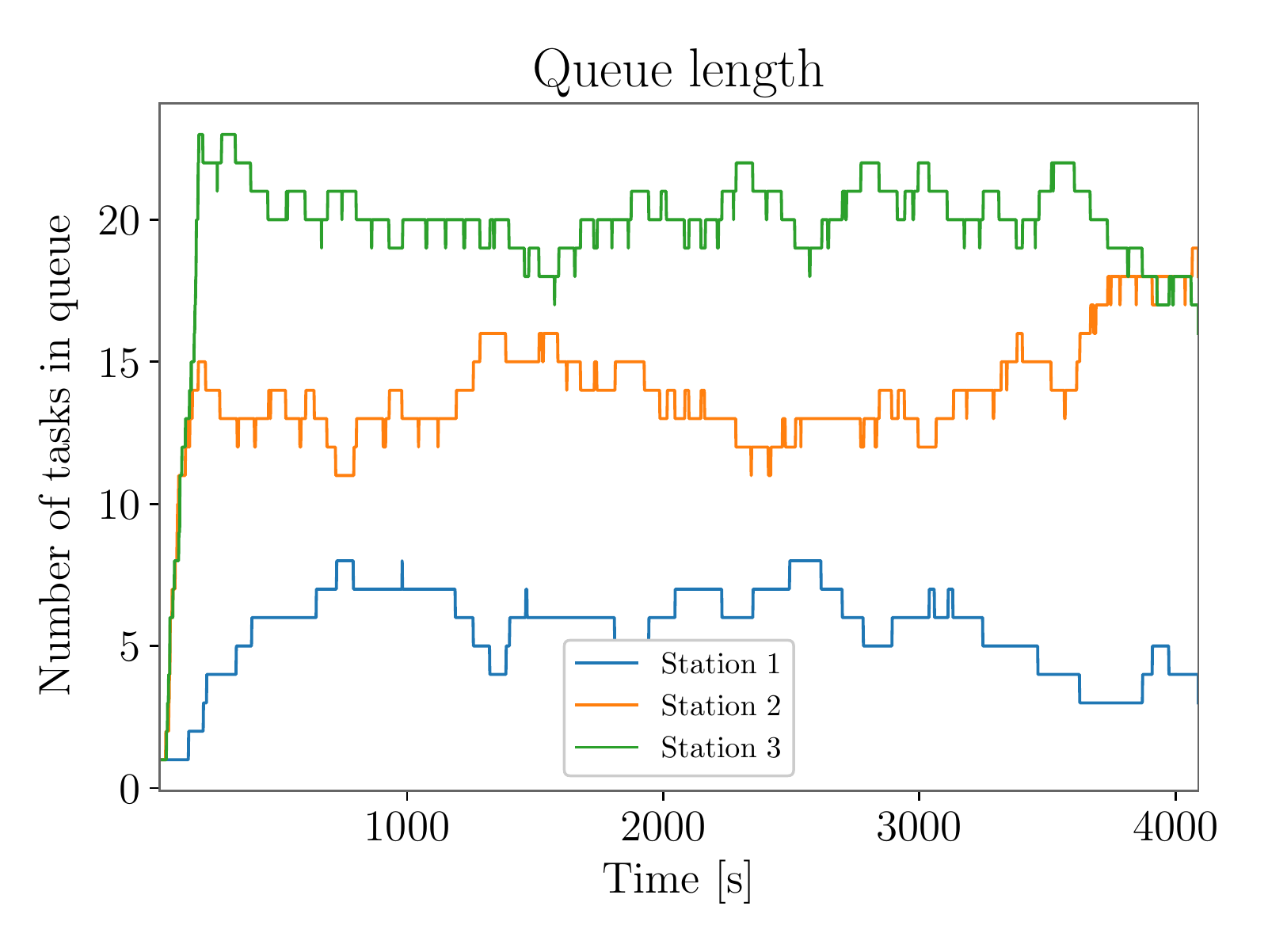}
        \caption{}
        \label{fig:scenario_5_all_figs:queue_length}
    \end{subfigure}
    \vfill
    \begin{subfigure}[t]{0.45\columnwidth}
        \centering
        \includegraphics[width = \columnwidth]{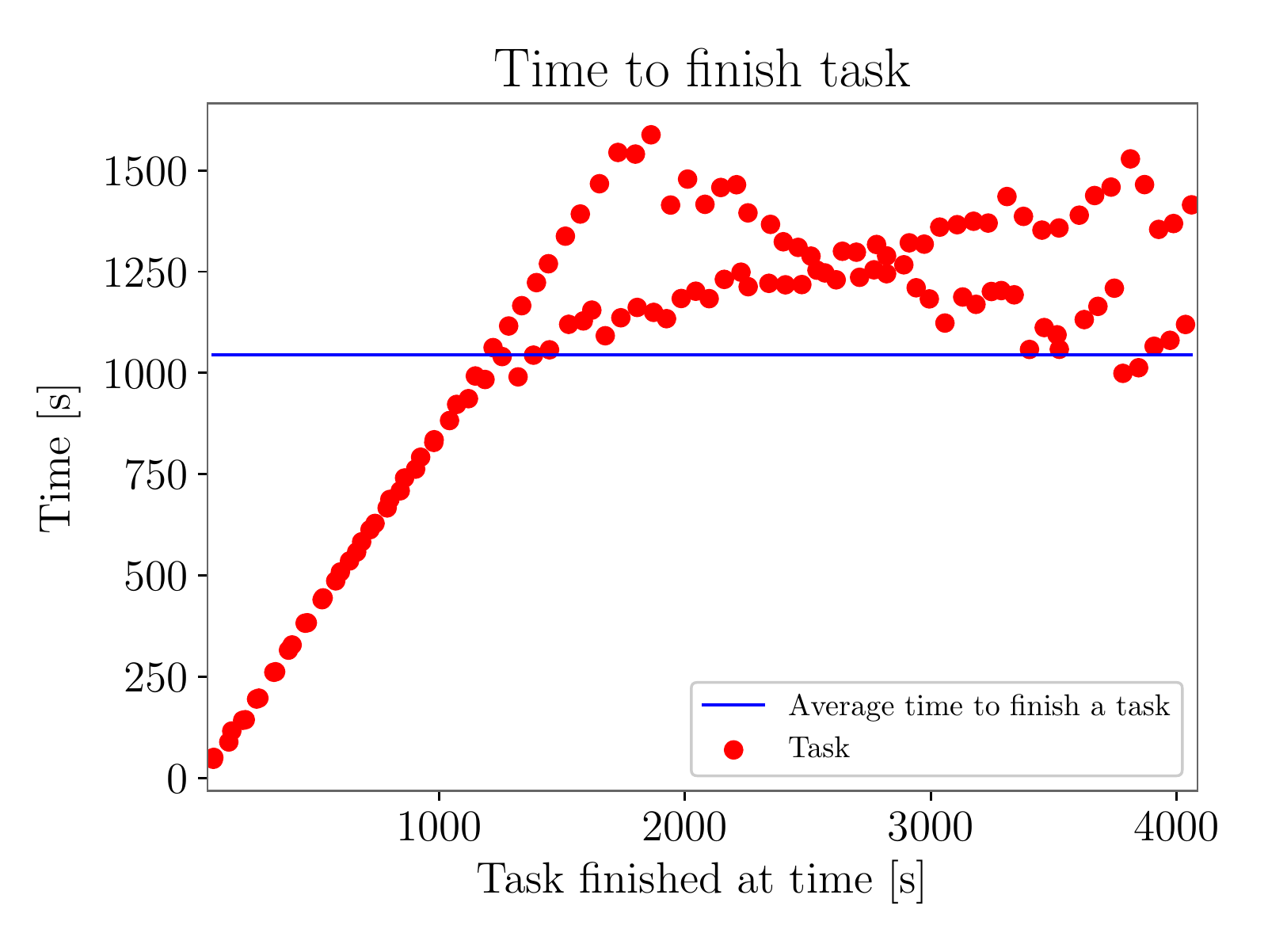}
        \caption{}
        \label{fig:scenario_5_all_figs:time_to_finish}
    \end{subfigure}
    \hfill
    \begin{subfigure}[t]{0.45\columnwidth}
        \centering
        \includegraphics[width = \columnwidth]{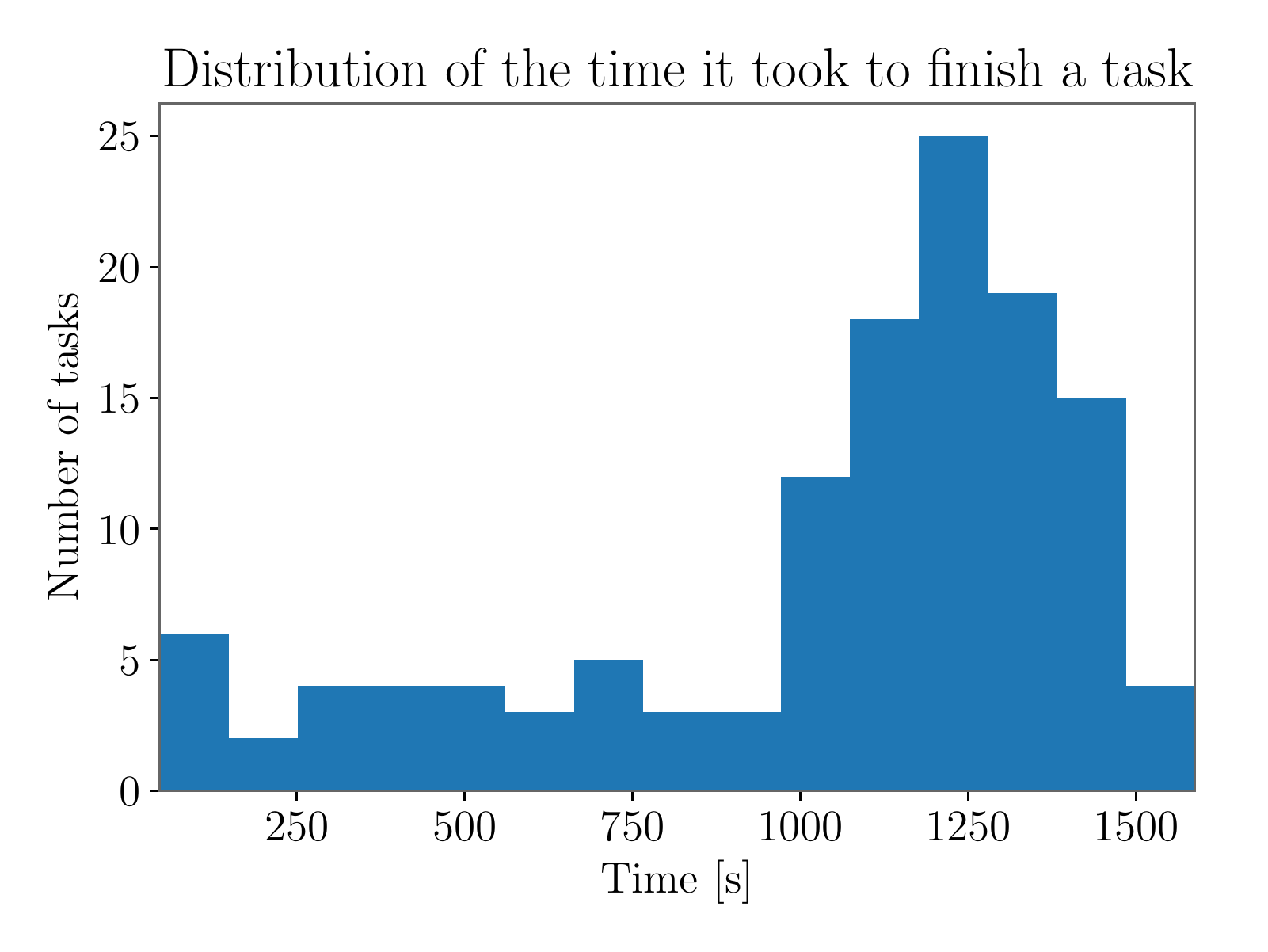}
        \caption{}
        \label{fig:scenario_5_all_figs:distribution}
    \end{subfigure}
    \caption{Summary of the results related to scenario 5. (a) The evolution of queue lengths during the evaluation scenario, (b) The time taken to complete the tasks after they have been added to the pool of available tasks and (c) A distribution of the times taken to complete the available tasks.}
    \label{fig:scenario_5_all_figs}
\end{figure}

In the fifth scenario three stations start with empty queues and two mobile agents are available for executing tasks. Every second there is a probability of \(5\%\), \(15\%\) and \(15\%\) for each, respective, queue to add a task. There is also a global maximum of the total number of tasks that can be available of \(40\) tasks, this means that if the maximum number of total tasks is reached, no more tasks can be added by any station. In this scenario the parameter \(q\) is adjusted to
\begin{equation}
    q = 0
\end{equation}
to demonstrate how the system behaves when not considering how long each queue is. The results can be seen in Fig \ref{fig:scenario_5_all_figs}. As expected the different queues will remain unbalanced, but, as can be seen in Fig. \ref{fig:scenario_5_all_figs:time_to_finish} and \ref{fig:scenario_5_all_figs:distribution} the waiting time of individual tasks becomes more stable.

\section{Discussion}

\subsection{Queue Lengths and Task Waiting Time}
The length of individual queues are well balanced in the static scenarios, shown in Fig. \ref{fig:scenario_1_queue_lengths} and Fig. \ref{fig:scenario_2_queue_lengths}, even though the individual tasks have different drop-off locations at various distances. In the dynamic scenario when the individual wait time of each task was not considered as a cost, shown in Fig \ref{fig:scenario_3_all_figs}, the queues are well balanced. Including the cost related to wait time of individual tasks is reducing the balancing between the queues, but every queue will still stabilize around a equilibrium point that depends on the relative inflow to the individual queues and the factors in the cost functions related to queue length and waiting time. These effects are visible if Fig. \ref{fig:scenario_3_all_figs:queue_length} and Fig. \ref{fig:scenario_4_all_figs:queue_length} are compared. The final scenario, shown in Fig \ref{fig:scenario_5_all_figs}, shows the results when no penalty is added based on queue length. Here it can be seen that the queues will be poorly balanced but the waiting time of individual tasks will become more stable.

Including the cost for individual tasks waiting time gives another interesting property, it is reducing the maximum waiting time of individual tasks and making sure that all tasks are serviced within a specific time. These results can be seen in Fig. \ref{fig:scenario_3_all_figs:time_to_finish}, \ref{fig:scenario_3_all_figs:distribution}, \ref{fig:scenario_4_all_figs:time_to_finish} and \ref{fig:scenario_4_all_figs:distribution}. The effects are especially clear if the distributions of waiting time are compared for scenario 3 and 4. It should also be noted that the average time to complete a task is not affected.

This gives the conclusion that the cost for queue length should be increased to prioritize balancing the lengths of the different queues and that increasing the cost related to waiting time of individual task leads to reducing the probability that some tasks has to wait much longer, compared to other tasks, to be serviced. It is also expected that the performance related to both the balancing and the wait time would improve if the constraint \(m_i\), meaning that only one agent can service a single queue at any time, was relaxed. This constraint, however, will be more and more relevant in bigger scenarios where it is only feasible to serve a queue with a maximum number of agents at the same time.

\subsection{Number of Agents and Stations}
A fundamental subject to analyse is what will happen if the number of agents or stations are increased. Here there are two possible cases, the first being that the total capacity of the agents is higher than the inflow of new tasks. In this case it is expected that all tasks will be completed as soon as they are added since there is an abundance of agents. The other case, being that the agents can not directly complete all the tasks that are introduced, is more interesting. Here, it is expected that a behavior similar to what is shown in Fig. \ref{fig:scenario_4_all_figs} where a steady state result is obtained. The resulting queue lengths, and the average waiting time of tasks, will depend on primarily the tuning parameters in the optimization problem but also on the specific tasks; meaning that if the tasks have different properties, such as drop-off location far from each other, the resulting behavior is expected to be more varying with time. The constraint that only \(m_i\) agents can be allocated to a queue should also be considered here; if the inflow of new tasks in a queue is greater than the capacity of \(m_i\) agents, that queue will no longer remain balanced but keep growing.

\section{Conclusions and Future Work}

We studied the problem of multi-agent pick up and delivery while also considering relative queue lengths, waiting time of individual tasks and constraints on how many agents can service a queue at the same time in a situation with no a priori knowledge of the tasks. To solve this problem we proposed a auction based framework to allocate and coordinate tasks to a team of mobile agents, this framework penalizes the cost to complete a task, the length of each queue and the waiting time of individual tasks. The different tuning parameters and their effects on the resulting behavior where studied in simulation using illustrative scenarios. These scenarios showed that the our approach for solving the specific problem are effective and able to handle various situation with an desirable outcome.

Future work includes analysing the computational complexity of both the optimization problem and the cost calculation by the agents, including investigating how to reduce the size of the resulting problem while still maintaining close to optimal solutions. 
To demonstrate the feasibility of this framework bigger simulation studies will be conducted to validate its scalability followed by real world experiments.

\bibliography{ifacconf}

\end{document}